\documentclass[11pt,a4paper]{article}
\usepackage[hyperref]{acl2021}
\usepackage{graphicx}
\usepackage{enumitem}
\usepackage{times}
\usepackage{latexsym}

\usepackage{microtype}

\usepackage{float}

\aclfinalcopy 
\def\aclpaperid{1539} 

\title{RAW-C: Relatedness of Ambiguous Words---in Context\\ (A New Lexical Resource for English)}

\author{Sean Trott \\
  University of California, San Diego \\
  \texttt{sttrott@ucsd.edu} \\\And
  Benjamin Bergen \\
  University of California, San Diego \\
  \texttt{bkbergen@ucsd.edu} \\}

\date{}

\begin{document}
\maketitle
\begin{abstract}
Most words are \textit{ambiguous}---they convey distinct meanings in different contexts---and even the meanings of unambiguous words are \textit{context-dependent}. Both phenomena present a challenge for NLP. Recently, the advent of contextualized word embeddings has led to success on tasks involving lexical ambiguity, such as Word Sense Disambiguation. However, there are few tasks that directly evaluate how well these embeddings accommodate the continuous, dynamic nature of word meaning---particularly in a way that matches human intuitions. We introduce \textbf{RAW-C}, a dataset of graded, human relatedness judgments for 112 ambiguous words in context (with 672 sentence pairs total), as well as human estimates of sense dominance. The average inter-annotator agreement for the relatedness norms (assessed using a leave-one-annotator-out method) was 0.79. We then show that a measure of cosine distance, computed using contextualized embeddings from BERT and ELMo, correlates with human judgments, but that cosine distance also systematically \textit{underestimates} how similar humans find uses of the same sense of a word to be, and systematically \textit{overestimates} how similar humans find uses of different-sense homonyms. Finally, we propose a synthesis between psycholinguistic theories of the mental lexicon and computational models of lexical semantics.
\end{abstract}

\section{Introduction}\label{sec:intro}

Words mean different things in different contexts. Sometimes these meanings appear to be distinct, a phenomenon known as \textit{lexical ambiguity}. In English, approximately 7\% of wordforms are \textit{homonymous}, i.e., they have multiple, unrelated meanings\footnote{\citet{dautriche2015weaving} estimates the average rate of homonymy across languages to be 4\%.} (e.g., ``tree bark'' vs. ``dog bark''), and as many as 84\% of wordforms are \textit{polysemous}, i.e., they have multiple, related meanings (e.g., ``pet chicken'' vs. ``roast chicken'') \cite{rodd2004modelling}. But even unambiguous words evoke subtly different interpretations depending on the context of use, i.e., their meanings are dynamic and \textit{context-dependent} \cite{yee2016putting, li2021word}. While the uses of \textit{runs} in ``the boy runs'' vs. ``the cheetah runs'' may not be considered distinct meanings, a human comprehender will likely activate a different mental image when processing each sentence \cite{elman2009meaning}. 

These facts present a challenge for computational models of lexical semantics. Any downstream task that involves meaning requires models capable of \textit{disambiguating} among the multiple possible meanings of an ambiguous word in a given context. Further, the \textit{graded} nature of human semantic representations can influence how comprehenders construe events and participants in those events \cite{elman2009meaning, li2021word}. In turn, a number of Natural Language Processing (NLP) tasks could benefit from context-sensitive representations that go beyond discrete sense representations and capture the manner in which humans construe events---including sentiment analysis, bias detection, machine translation, and more \cite{trott2020re}. If an eventual goal of NLP is human-like language understanding, models must be equipped with semantic representations that are flexible enough to accommodate the dynamic, context-dependent nature of word meaning---as humans appear to do \cite{elman2009meaning, li2021word}. 

Yet a crucial prerequisite to developing better models is \textit{evaluating} those models along the relevant dimensions of performance. Thus, at the minimum, we need metrics that evaluate a model along two critical dimensions:

\begin{enumerate}[itemsep=-1ex]
    \item \textbf{Disambiguation}: A model's ability to distinguish between distinct meanings of a word.\label{criteria1}
    \item \textbf{Contextual Gradation}: A model's ability to modulate a given meaning in context, in ways that reflect the continuous nature of human judgments.\label{criteria2}
\end{enumerate}

A promising development in recent years is the rise of contextualized word embeddings, produced using neural language models such as BERT \cite{devlin2018bert}, ELMo \cite{peters2018deep}, XLNet \cite{yang2019xlnet}, and more. Advances in these models have yielded improved performance on a number of tasks, including Word Sense Disambiguation (WSD) \cite{boleda2019putting, loureiro2020language}.

WSD satisfies the Disambiguation Criterion above, but not the Contextual Gradation Criterion. In fact, there is still a dearth of metrics for assessing the degree to which contextualized representations match human judgments about the way in which context shapes meaning.

In Section \ref{sec:related}, we describe several related datasets that satisfy at least one of these criteria. In Section \ref{sec:dataset}, we introduce and describe the dataset construction process for RAW-C: Relatedness of Ambiguous Words---in Context.\footnote{The dataset can be found on GitHub: \url{https://github.com/seantrott/raw-c}.} In Section \ref{sec:annotation}, we describe the procedure we followed for collecting human relatedness norms for each sentence pair. In Section \ref{sec:analysis}, we report the results of several analyses that probe how well contextualized embeddings from two neural language models (BERT and ELMo) predict these norms. Finally, in Section \ref{sec:discussion}, we explore possible shortcomings in current models, and propose avenues for future work.

\section{Related Work}\label{sec:related}

Most existing datasets fulfill either the \textbf{Disambiguation} or the \textbf{Contextual Gradation} criterion, but few datasets fulfill both (see \citet{haber2020assessing} for an exception).

Several datasets contain human relatedness and similarity judgments for \textit{distinct} words in isolation (see Section \ref{subsec:simrel}). Others are used for Word Sense Disambiguation, and contain ambiguous words in different sentence contexts, along with annotated sense labels (see Section \ref{subsec:wsd}); as noted in the Introduction, WSD fulfills the Disambiguation Criterion, but not the Contextual Gradation Criterion. Several recent datasets contain graded relatedness judgments for words in different contexts (see Section \ref{subsec:graded}). However, none focus specifically on graded relatedness judgments for \textit{ambiguous} words, controlling both the inflection and part of speech of the target word in question. Finally, one dataset \cite{haber2020assessing} contains similarity judgments for polysemous words in context, but is more limited in size and does not match the sentence frame across the two uses (see Section \ref{subsec:graded_ambig}).

\subsection{De-contextualized Word Similarity and Relatedness}\label{subsec:simrel}

Several datasets contain human judgments of the \textit{similarity} or \textit{relatedness} of (mostly English) word pairs, in isolation (see \citet{taieb2020survey} for a review). This includes SimLex-999 \cite{hill2015simlex}, SimVerb-3500 \cite{gerz2016simverb}, WordSim-353 \cite{finkelstein2001placing}, MTurk-771 \cite{halawi2012large}, MEN \cite{bruni2014multimodal}, and more. These datasets are primarily used for evaluating the quality of static semantic representations, including distributed semantic models such as GloVe \cite{pennington2014glove}, as well as representations that use knowledge bases like WordNet \cite{faruqui2015non}. 

However, these resources are (by definition, as decontextualized judgments) not directly amenable to evaluating how well a model incorporates \textit{context} into its semantic representation of a given word.  
\subsection{Word Sense Disambiguation}\label{subsec:wsd}

In Word Sense Disambiguation (WSD), a classifier predicts the ``sense'' of an ambiguous word in a given context, often using a contextualized embedding. WSD relies on annotated sense labels, which in turn requires determining whether any given pair of word uses belong to the same or distinct senses---i.e., whether to ``lump'' or ``split''. There is considerable debate about how \textit{granular} word sense inventories should be \cite{hanks2000word, brown2008choosing};\footnote{This also raises deeper philosophical issues about exactly what qualifies as a ``sense'' \cite{hanks2000word, tuggy1993ambiguity, geeraerts1993vagueness, kilgarriff2007word}; answering these questions is beyond the scope of this paper, though see Section \ref{sec:discussion} for a brief discussion.} resources range in granularity from WordNet \cite{fellbaum1998} to the Coarse Sense Inventory, or CSI \cite{lacerra2020csi}. Recent work using coarse-grained sense inventories has achieved success rates of 85\% and beyond \cite{lacerra2020csi, loureiro2020language}. 

In terms of the criteria listed above, WSD satisfies the Disambiguation Criterion, but not the Contextual Gradation Criterion. WSD only captures a model's ability to distinguish between distinct senses; it does not assess how meaning is modulated within a given sense category, e.g., that a human comprehender might consider the meaning of \textit{runs} in ``the cheetah runs'' as more similar to ``the jaguar runs'' than to ``the toddler runs'', or that some uses of a sense might be more prototypical than others.

\subsection{Contextualized Word Similarity and Relatedness}\label{subsec:graded}

There have been several recent efforts to address this gap in the literature:

The Stanford Contextual Word Similarity (SCWS) dataset \cite{huang2012improving} contains similarity judgments for 2,003 English word pairs in a sentence context. Approximately 12\% of the pairs contain the same word (e.g., ``pack his bags'' vs. ``pack of zombies''), though not always in the same part of speech; in most cases, the words compared are different (e.g., ``left'' vs. ``abandon''). This dataset is a useful step towards contextualized similarity judgments, but because most pairs contain different words (or the same word in different parts of speech), static word embeddings such as Word2Vec can still perform quite well without considering the context at all \cite{pilehvar2018wic}. 

The Word in Context (WiC) dataset \cite{pilehvar2018wic} contains a set of over 7,000 sentence pairs with an overlapping English word, labeled according to the use of that word corresponds to same or different senses. As \citet{pilehvar2018wic} note, the structure of the dataset requires some form of contextualized meaning representation to perform above a random baseline, which makes it more suitable for interrogating contextualized embeddings. However, the task is a binary classification task along the lines of WSD, making it harder to assess the Contextual Gradation Criterion.

The CoSimLex dataset \cite{armendariz2020semeval}, created with the Graded Word Similarity in Context (GWSC) task, contains graded similarity judgments for a number of word pairs across English (340), Croatian (112), Slovene (111), and Finnish (24). Each pair of words is rated in two separate contexts, yielding 1174 scores in total. Sentence contexts were extracted from each language's Wikipedia. Unlike WiC, the word pairs do not actually contain the same word---rather, for any given word pair (e.g., ``beach'' and ``seashore''), there are at least two pairs of sentence contexts with associated similarity judgments. Thus, this dataset can be used to assess graded differences in contextualized meaning representations, but not directly for the \textit{same} ambiguous word. 

\subsection{Contextualized Similarity of Ambiguous Words}\label{subsec:graded_ambig}

Finally, one dataset \cite{haber2020assessing, haber2020word} contains graded similarity judgments (as well as co-predication acceptability judgments) for a number of polysemous words in distinct sentential contexts, meeting both \textbf{Contextual Gradation} and the \textbf{Disambiguation} criteria. 

The main limitations of this dataset are its size (it contains examples for only 10 polysemes), as well as the fact that the sentence frames are also not always controlled for each polysemous word. 

\subsection{Summary}

Most datasets reviewed above allow practitioners to evaluate models on their ability to disambiguate (i.e., the Disambiguation Criterion) or their ability to capture graded differences in word relatedness (i.e., the Contextual Gradation Criterion); one dataset \cite{haber2020assessing, haber2020word} meets both criteria.

But to our knowledge, no datasets contain graded relatedness judgments for ambiguous words in tightly controlled sentence contexts, with inflection and part-of-speech controlled across each use. In Section \ref{sec:dataset} below, we describe the procedure we followed for constructing such a dataset.

\section{RAW-C: Relatedness of Ambiguous Words, in Context}\label{sec:dataset}

Items were adapted from stimuli used in past psycholinguistic studies, which contrasted behavioral responses to homonymous and polysemous words, either in isolated lexical decision tasks \cite{klepousniotou2007disambiguating} or in a disambiguating context \cite{klepousniotou2002processing, klepousniotou2008making, brown2008polysemy}. We selected 115 words in total. For each ambiguous word (e.g., ``bat''), we created four sentences: two each for two distinct meanings of the word. We attempted to match the sentence frames as closely as possible, in most cases altering only a single word\footnote{There were 13 words for which at least one of the four sentences used a different article (``a'' vs. ``an''), in addition to having a different disambiguating word.} across the four sentences to disambiguate the intended meaning:

\begin{enumerate}[itemsep=-1ex]
    \item[1a.] He saw a \textit{fruit} bat.
    \item[1b.] He saw a \textit{furry} bat. 
    \item[2a.] He saw a \textit{wooden} bat.
    \item[2b.] He saw a \textit{baseball} bat.
\end{enumerate}

We also labeled each word according to whether the two distinct meanings were judged by lexicographers to be Polysemous or Homonymous. Distinguishing homonymy from polysemy is notoriously challenging \cite{valera2020polysemy}; common tests include determining whether the two meanings share an etymology (polysemy) or not (homonymy), or determining whether the two meanings are conceptually related (polysemy) or not (homonymy). Both tests can be criticized on multiple grounds \cite{tuggy1993ambiguity, valera2020polysemy}, and do not always point in the same direction (e.g., etymologically related words sometimes drift apart, resulting in apparent homonymy).

For our annotation, we consulted both the online Merriam-Webster Dictionary (\url{https://www.merriam-webster.com/}) and the Oxford English Dictionary, or OED (\url{https://www.oed.com/}), and identified whether each dictionary listed the two meanings in question in separate lexical entries (homonymy), or as different senses under the same lexical entry (polysemy).\footnote{Our primary goal with this labelling was not to definitively distinguish homonymy from polysemy; as noted above, there is no single, universal criterion for doing so, and different criteria might be more or less relevant for different purposes. It was simply to specify how lexicographers treat the different words, in case that information is useful for users of the resource.} For example, both dictionaries list the \textit{animal} and \textit{meat} senses of the word ``lamb'' as different senses under the same lexical entry, whereas they list the \textit{animal} and \textit{artifact} senses of the word ``bat'' under different lexical entries. There was one word (``drill'') on which the two dictionaries did not agree; in this case, we labeled the two meanings (``electric drill'' vs. ``grueling drill'') as homonymy (as per the OED).

There were also three words for which neither dictionary distinguished the two meanings (either in terms of homonymy or polysemy). For example, ``best-selling novel'' and ``thick novel'' refer to \textit{cultural} and \textit{physical} artifacts, respectively, but are not listed as distinct senses. Again, this highlights the challenge of distinguishing outright \textit{ambiguity} from \textit{context-dependence}; these items were included in the annotation study described below, but were excluded from the final set of norms, thus resulting in 112 target words altogether.\footnote{The existence of these ``Unsure'' items, as well as items for which the two dictionaries disagreed on the issue of homonymy vs. polysemy, raises the question of whether empirical measurements such as relatedness judgments (or even cosine distance) could help inform lexicographic decisions. As a proof of concept, we trained a logistic regression classifier (using leave-one-out cross-validation) to predict whether two contexts of use belonged to the Same Sense, using Mean Relatedness. The classifier successfully categorized 86.76\% of held-out test items as belonging to the same or different senses. Further, for different sense items only, a trained classifier successfully categorized 79\% of held-out test items as polysemous or homonymous. While only a proof of concept, this demonstration suggests a promising avenue for future research.} Each word was used in four sentences, for a total of six \textit{sentence pairs} (see Table \ref{table:descriptive} for more details). 84 of the target words were nouns, and 28 were verbs (note that Part-of-Speech was always held constant \textit{within} each word).

\begin{table}[H]
\begin{tabular}{lll}
\hline
Ambiguity Type & \#Words & \#Sentence Pairs \\
\hline
Homonymy       & 38      & 228              \\
Polysemy       & 74      & 444             
\end{tabular}
\caption{Number of words (and sentence pairs) for each type of ambiguity.}\label{tab:accents}
\label{table:descriptive}
\end{table}

\section{Human Annotation}\label{sec:annotation}

\subsection{Participants}

81 participants were recruited through UC San Diego's undergraduate subject pool for Psychology, Cognitive Science, and Linguistics students. Participants received class credit for participation. Three participants were removed for failing the bot checks at the beginning of the study, and one was removed for failing the catch trials embedded in the experiment, leaving 77 participants in total (59 Female, 16 Male, 2 Non-binary). The median age of participants was 20 (M = 20.22, SD = 2.7), with ages ranging from 18 to 38. 74 participants self-reported as being native speakers of English.

\subsection{Materials}

We used the original set of 115 words described in Section \ref{sec:dataset}, i.e., including the three items labeled ``Unsure''. Each word had four sentences; accounting for order, this resulted in twelve possible sentence pairs (six pairs, with two orders each) for each word, for a total 1380 items.

\subsection{Procedure}

After giving consent, participants answered two questions designed to filter out bots (e.g., ``Which of the following is not a place to swim?'', with the correct answer being ``Chair''). They were then given instructions, which included a description of how the meaning of a word can change in different contexts.

\begin{figure}[H]
    \centering
    \includegraphics[width=7cm]{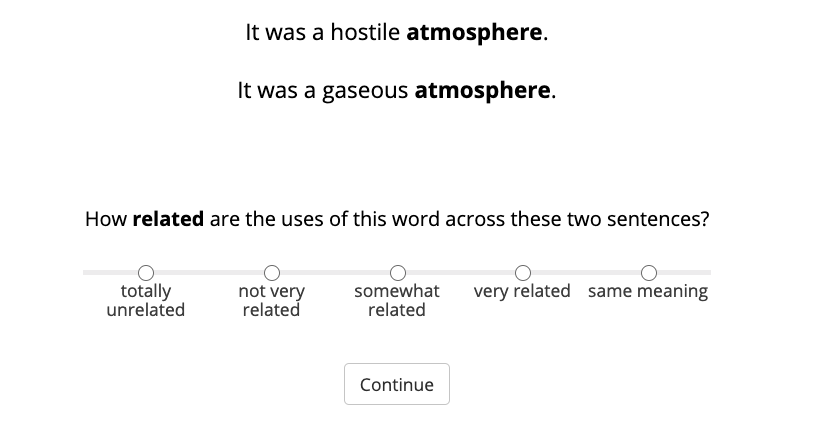}
    \caption{Example item from study.}
    \label{fig:study}
\end{figure}

On each page of the study, participants were shown a pair of sentences, with the target word bolded (see Figure \ref{fig:study} for an example). They were asked to indicate how \textbf{related} the uses of that word were across the two sentences, with a labeled Likert scale ranging from ``totally unrelated'' to ``same meaning''.

We included two ``catch'' trials in the study to identify participants who did not pay attention. In one, the two sentences were identical, such that the correct answer is ``same meaning''; the other featured a homonym with two different parts of speech (\textit{rose.v} and \textit{rose.n}), such that the correct answer was ``totally unrelated''.

Excluding the catch trials, participants saw 115 sentence pairs total; no word was repeated twice across trials for the same participant. The comparisons any given subject saw for a given word were randomly sampled from the 12 possible sentence pairs, and the order of trials was randomized.\footnote{Based on the suggestion of an anonymous reviewer, we also ran a follow-up norming study to collect estimates of sense frequency bias (sometimes called \textit{dominance}); sense dominance is known to play an important role in the processing of ambiguous words \cite{klepousniotou2007disambiguating, rayner1994effects, binder1998contextual, leinenger2013eye}. These dominance norms are included in the final dataset.}

\section{Analysis and Results}\label{sec:analysis}

The analyses run below were performed on the 112 target words (i.e., excluding the ``Unsure'' items). Human annotations were assigned to a scale from 0 (``totally unrelated'') to 4 (``same meaning''). 

\subsection{Analysis of Sentence Pairs}

Before analyzing the responses of human annotators, we first sought to characterize how well two neural language models captured the \textit{categorical} structure in the dataset---i.e., whether their contextualized representations could be used to distinguish same-sense from different-sense uses of the same word, as well as words labelled as different-sense Homonyms from different-sense Polysemes.

\begin{figure}[H]
    \centering
    \includegraphics[width=7cm]{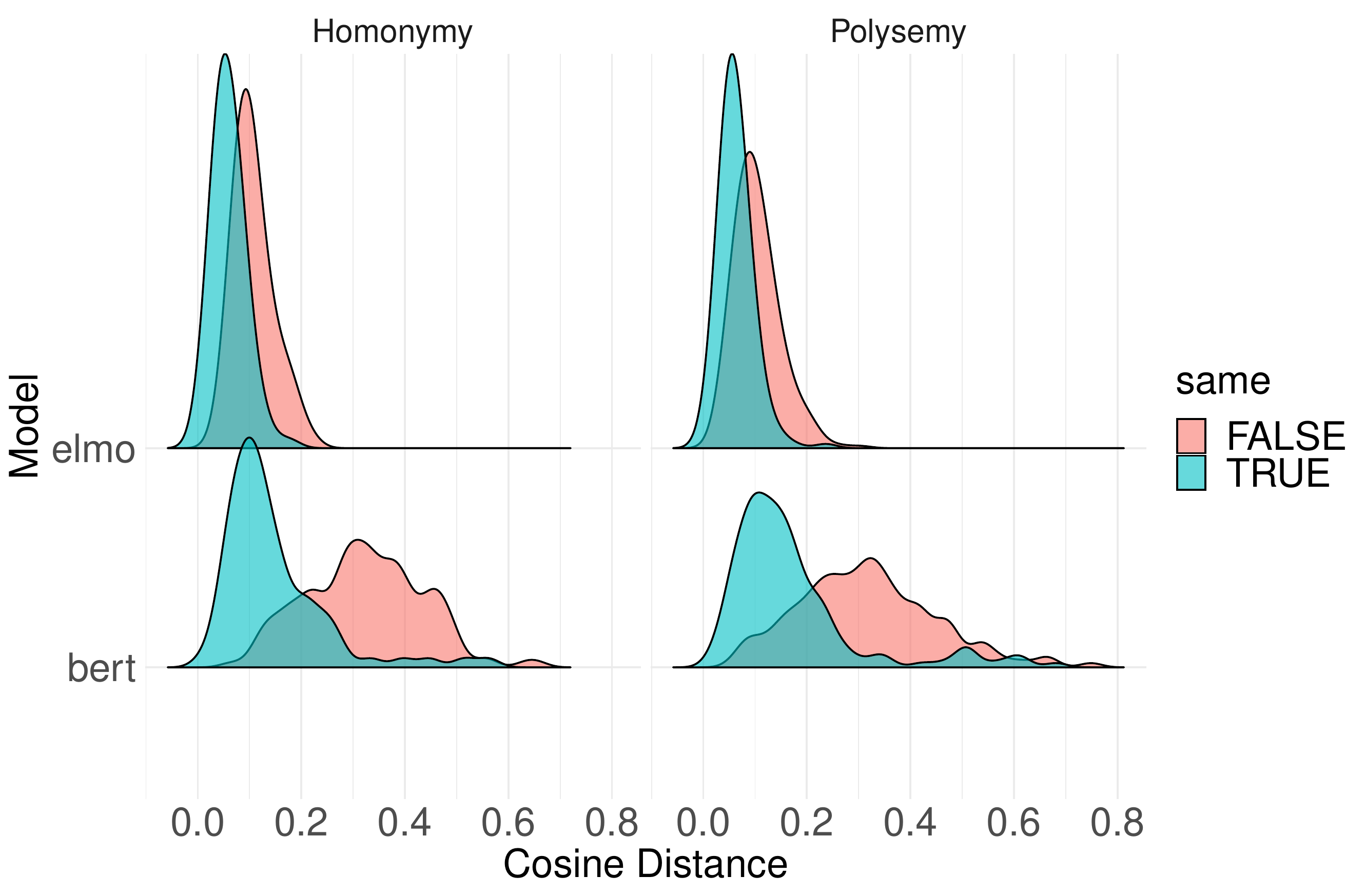}
    \caption{Cosine Distances between the target word's contextualized embeddings for both language models, plotted by Same Sense (True vs. False) and Ambiguity Type (Homonymy vs. Polysemy).}
    \label{fig:cosines}
\end{figure}

We ran every sentence through two language models: ELMo, using the Python AllenNLP package \cite{Gardner2017AllenNLP}, and BERT, using the \texttt{bert-embedding} package.\footnote{\url{https://pypi.org/project/bert-embedding/}} Then, for each sentence pair, we computed the Cosine Distance between the contextualized representations of the target wordform (e.g., \textit{bat} in ``He saw the furry bat'' and ``He saw the wooden bat''). The distribution of Cosine Distances is visualized in Figure \ref{fig:cosines}. 

We also performed several statistical analyses, using the \texttt{lme4} package in R \cite{bates2007lme4}. In each case, we compared a full model to a reduced model using a log-likelihood ratio test. All models had Cosine Distance as a dependent variable, and included Part-of-Speech as a fixed effect, random intercepts for Word and Language Model (i.e., ELMo vs. BERT), and by-Word random slopes for the effect of Same Sense.

Adding a fixed effect of Same Sense significantly improved model fit $[{\chi}{^2}(1)=143.72, p < .001]$, with same-sense uses significantly closer than different-sense uses $[\beta=-.099, SE = 0.005]$. However, adding an interaction between Same Sense and Ambiguity Type (as well as fixed effects of both) did not significantly improve the fit above a model omitting the interaction $[{\chi}{^2}(1)=2.19, p = 0.14]$. In other words, both language models could differentiate same-sense and different-sense uses of an ambiguous word, but their ability to discriminate between Homonymy and Polysemy was marginal at best.

\subsection{Analysis of Human Annotations}\label{subsec:human_analysis}

Our primary goal was understanding the distribution of human relatedness annotations---both in terms of how it reflects the underlying categorical structure of the dataset (e.g., Homonymy vs. Polysemy), as well as the Cosine Distance measures from each language model. As in the section above, we constructed a series of linear mixed effects models and performed log-likelihood ratio tests for each model comparison; in each case, the dependent variable was Relatedness. All models included a fixed effect of Part-of-Speech, by-subject and by-word random slopes for the effect of Same Sense, by-subject random slopes for the effect of Ambiguity Type, and random intercepts for subjects and items.

\begin{figure}[H]
    \centering
    \includegraphics[width=7cm]{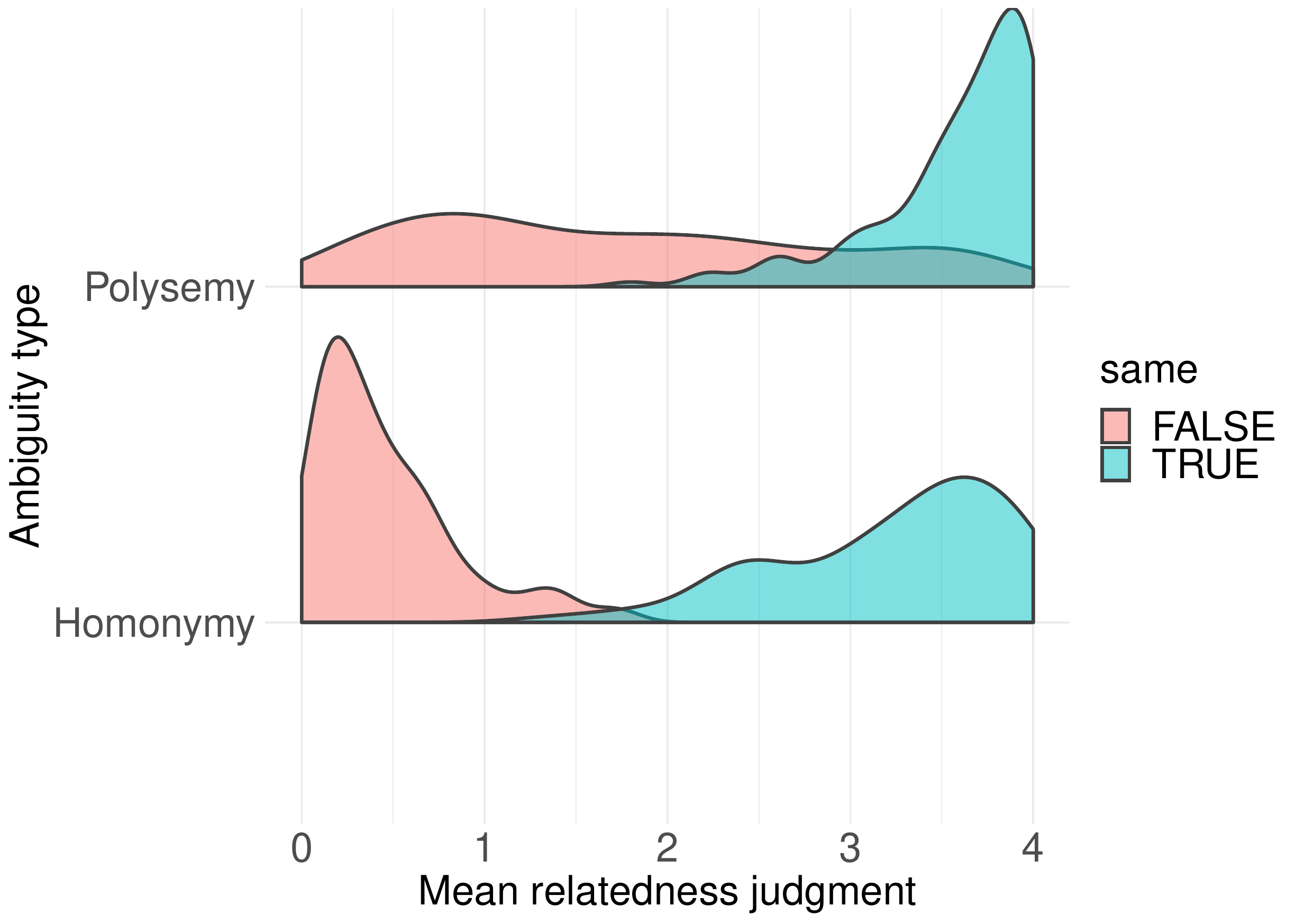}
    \caption{Mean relatedness judgments for each sentence pair, plotted by by Same Sense (True vs. False) and Ambiguity Type (Homonymy vs. Polysemy).}
    \label{fig:relatedness}
\end{figure}

First, we asked whether participants' relatedness judgments varied across same-sense and different-sense sentence pairs. We added a fixed effect of Same Sense to the base model described above, along with fixed effects for the Cosine Distance measures from BERT and ELMo. This model explained significantly more variance than a model omitting only Same Sense $[{\chi}{^2}(1)=207.11, p < .001]$, with same-sense uses receiving higher relatedness judgments on average $[\beta=1.94, SE = 0.1]$. The median relatedness judgment for same-sense uses was 4 ($M = 3.46, SD = 1.02$), while the median relatedness judgment for different-sense uses was 1 ($M = 1.31, SD = 1.45$). 
Second, we asked whether participants' judgments were sensitive to the distinction between Homonymy and Polysemy. We added an interaction between Same Sense and Ambiguity Type (along with a fixed effect of Ambiguity Type) to the model described above. The interaction significantly improved model fit $[{\chi}{^2}(1)=25.45, p < .001]$. The median relatedness for both same-sense homonyms and polysemes was 4, whereas the median relatedness for different-sense homonyms (0) was lower than that for different-sense polysemes (2). Further, as depicted in Figure \ref{fig:relatedness}, there was considerably more variance across polysemous words than homonymous words---this makes sense, given that some polysemous meanings are highly related (e.g., ``pet chicken'' vs. ``roast chicken''), while others are more distant (e.g., ``desperate act'' vs. ``magic act''). 

Third, we asked whether the Cosine Distance measures explained independent variance above and beyond that explained by Same Sense and Ambiguity Type. A full model including all factors explained more variance than a model excluding only the Cosine Distance measure from BERT $[{\chi}{^2}(1)=36.19, p < .001]$, as well as a model excluding only the Cosine Distance measure from ELMo $[{\chi}{^2}(1)=16.92, p < .001]$. This indicates that Relatedness does not vary purely as a function of the categorical structure in the dataset---the graded relatedness judgments were sensitive to subtle differences in context.

\subsection{Inter-Annotator Agreement}

Inter-annotator agreement was assessed by calculating the average Spearman's rank correlation between each participant's responses and the Mean Relatedness for the set of 112 items observed by that participant---where Mean Relatedness was calculated after omitting responses by the participant in question. This answers the question: to what extent do each participant's responses correlate with the consensus rating by the 76 other participants? Using this method, the average correlation was $\rho = 0.79$, with a median of $\rho=0.81$ ($SD = .07$). The lowest agreement was $\rho = 0.55$, and the highest was $\rho = 0.88$. 

\subsection{Evaluation of Language Models}\label{evaluation}

To evaluate the language models, we collapsed across the single-trial data and computed the Mean and Median Relatedness for each unique sentence pair. The distribution of Mean Relatedness judgments is depicted in Figure \ref{fig:relatedness}.

As in past work \cite{hill2015simlex}, we computed the Spearman's rank correlation between the distribution of Cosine Distance measures (from each model) and the Mean Relatedness for a given sentence pair. BERT performed slightly better than ELMo ($BERT_{\rho} = -0.58, ELMo_{\rho} = -0.53$).\footnote{Note that larger values of Cosine Distance indicate a \textit{larger} distance between two vectors; thus, a negative correlation is expected between relatedness and Cosine Distance.} Putting this in context, both models performed considerably worse than the average inter-annotator agreement score ($\rho = 0.79$). 

We also computed the $R^2$ of a linear regression including the Cosine Distance measures from \textit{both} BERT and ELMo. Combined, both measures explained roughly 37\% of the variance in Mean Relatedness judgments ($R^2 = 0.37$). Surprisingly, this was only slightly more than half the variance explained by a linear regression including only the interaction between Same Sense and Ambiguity Type ($R^2 = 0.66$), as well as a regression including all factors ($R^2 = 0.71$). 

\begin{figure}[H]
    \centering
    \includegraphics[width=7cm]{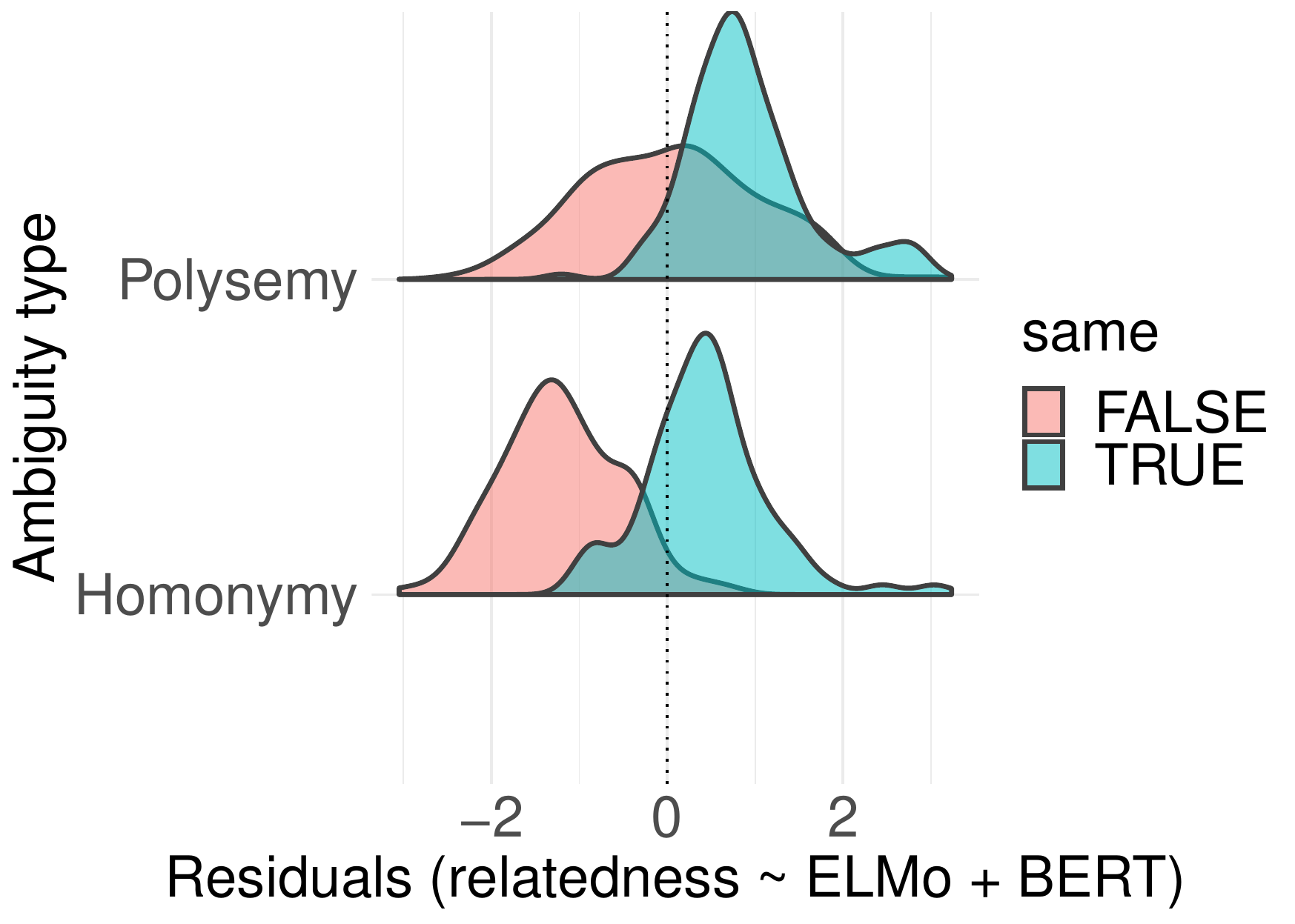}
    \caption{Residuals of a linear regression including Cosine Distance measures from both BERT and ELMo, plotted by by Same Sense (True vs. False) and Ambiguity Type (Homonymy vs. Polysemy).}
    \label{fig:residuals}
\end{figure}

By visualizing the residuals from the linear regression with only BERT and ELMo (see Figure \ref{fig:residuals}), we see that Cosine Distance appears to systematically \textit{underestimate} how related participants find same-sense judgments to be (for both Polysemy and Homonymy). Further, we see that Cosine Distance systematically \textit{overestimates} how related participants find different-sense Homonyms to be.

\section{Discussion}\label{sec:discussion}

Word meanings are dynamic, dependent on the contexts in which those words appear---and some words are even ambiguous, generating distinct, incompatible interpretations in different situations (e.g., ``fruit \textit{bat}'' vs. ``baseball \textit{bat}''). 

RAW-C contains graded relatedness judgments (by human annotators) for ambiguous English words in distinct sentential contexts. Importantly, the ambiguous wordform (e.g., ``bat'') is always matched for both part-of-speech and inflection across each sentence pair; 84 of the target words are nouns, and 28 are verbs. Each word has relatedness judgments for six different sentences pairs (four unique sentences): two same-sense pairs, and four different-sense pairs. Same sense pairs convey the same meaning, according to Merriam-Webster and the OED (e.g., ``fruit bat'' and ``furry bat''), while different sense pairs correspond to meanings listed in either distinct lexical entries (e.g., ``fruit bat'' and ``wooden bat'') or distinct sub-entries (e.g., ``marinated lamb'' and ``baby lamb''). Furthermore, different-sense pairs are labeled according to whether they are related via homonymy or polysemy, a relevant distinction for both lexicographers and psycholinguists---recent evidence suggests that polysemous and homonymous meanings are represented differently in the mental lexicon \cite{klepousniotou2002processing, klepousniotou2007disambiguating}. Finally, the sentential context is always tightly controlled; in most pairs, only one word differs across the two sentences (e.g., ``fruit'' vs. ``furry''). 

In Section \ref{sec:analysis}, we reported several primary findings. First, contextualized representations from both BERT and ELMo capture the distinction between same-sense and different-sense uses of a word, but their ability to distinguish between homonymy and polysemy is marginal at best. This contrasts with other recent work \cite{nair2020contextualized}, suggesting that BERT \textit{is} able to differentiate between homonymy and polysemy. One possible explanation for this difference in results is that \citet{nair2020contextualized} used naturally-occurring sentences from Semcor \cite{miller1993semantic}, whereas our sentence contexts were more tightly controlled. Our results indicate that even the presence of a single disambiguating word can trigger nuanced differences in semantic representation in humans, but not necessarily in current neural language models.

Second, we found that both BERT and ELMo explain independent sources of variance in human relatedness judgments, above and beyond Same Sense and Ambiguity Type (i.e., homonymy vs. polysemy). This is encouraging, because it demonstrates a direct benefit of graded (rather than categorical) judgments; for example, among the broad category of different-sense polysemous pairs, some are closely related (e.g., ``marinated lamb'' and ``baby lamb''), and others are considerably less closely related (e.g., ``hostile atmosphere'' and ``gaseous atmosphere''). Overall, contextualized embeddings from BERT were better at predicting human relatedness judgments than those from ELMo---this is consistent with past work \cite{wiedemann2019does} suggesting that BERT outperforms ELMo on tasks involving sense disambiguation. 

Importantly, however, both BERT and ELMo failed to capture variance in relatedness judgments that is captured by Same Sense and Ambiguity Type. As depicted in Figure \ref{fig:residuals}, Cosine Distance tended to \textit{underestimate} how related humans find same-sense uses to be, and \textit{overestimate} how related humans find different-senses to be. This is not entirely surprising, given that neither BERT nor ELMo are equipped with discrete sense representations---at most, they produce contextualized embeddings that are amenable to supervised classification or unsupervised clustering. Yet this also illustrates that---at least on this task---humans \textit{do} appear to draw on some manner of (likely fuzzy) categorical representation, such that the difference between two contexts of use is \textit{compressed} for same-sense meanings, and \textit{exaggerated} for different-sense meanings (particularly for homonyms). This suggests several exciting avenues for future work: can neural language models such as BERT be augmented with semantic knowledge or representational schemes that improve their performance on RAW-C or similar tasks? Both possibilities are explored in Section \ref{subsec:future} below.

\subsection{Future Work}\label{subsec:future}

As \citet{bender2020climbing} note, most language models are trained on linguistic form alone. In contrast, human language knowledge is \textit{grounded} in our embodied experience of the world \cite{bisk2020experience}. To the extent that human sense representations are guided by distinct sensorimotor or social-interactional associations, equipping language models with this information ought to facilitate their ability to distinguish between distinct meanings of a word (i.e., the Disambiguation Criterion) and modulate a given meaning in context (i.e., the Contextual Gradation Criterion).  

Practitioners could also look to (and in turn, inform) models of the human mental lexicon \cite{nair2020contextualized}. Several promising models attempt to address the \textit{continuous} nature of word meaning, as well as the issue of apparent category boundaries (i.e., word senses) \cite{rodd2004modelling, elman2009meaning}; at this stage, the role of continuity vs. categorical structure in human sense representations remains an open question. Models such as SenseBERT \cite{levine-etal-2020-sensebert} incorporate high-level sense knowledge into internal representations from the beginning, and find improvements on several WSD tasks---would this approach, or others like it, yield an improvement on RAW-C as well?

\subsection{Limitations of Dataset}

RAW-C has multiple limitations, some of which could also be addressed in future work. First, the broad category of ``polysemy'' is often subdivided into different mechanisms or manners of conceptual relation, such as metaphor and metonymy. This distinction is also believed to be cognitively relevant, with some evidence that metaphorically related senses are represented differently than metonymically related ones \cite{klepousniotou2002processing, klepousniotou2007disambiguating, lopukhina2018mental, yurchenko2020metaphor}. Future work could annotate polysemous word pairs for whether they are related by metaphor, metonymy, or another class of semantic relation---annotations could even be made as granular as the specific semantic relation involved (e.g., Animal for Meat) \cite{srinivasan2015concepts}. This finer-grained coding could be used to assess exactly which kinds of semantic relation correlate with the distributional profile of word tokens---i.e., are accessible from linguistic form alone---and which require some external module, whether in the form of grounded world knowledge or a structured knowledge base.

Another possible limitation is the fact that RAW-C contains experimentally controlled minimal pairs, instead of naturally-occurring sentences \cite{nair2020contextualized, haber2020assessing, haber2020word}. On the one hand, naturalistic sentences are useful for evaluating models on WSD ``in the wild'' (and indeed, there are a number of useful datasets for this purpose; see Section \ref{sec:related}). On the other hand, controlled datasets are useful if one's goal is to better understand a particular model or linguistic phenomenon---especially if this also allows a direct comparison with human annotations. For example, our analyses suggest that human sense representations must involve some additional levels of processing or information beyond the statistical regularities in word co-occurrence captured by BERT and ELMo. Moving forward, we hope that experimentally controlled datasets such as RAW-C will serve as a useful complement to existing, more naturalistic datasets.

\section{Conclusion}

We have presented a novel dataset for evaluating contextualized language models: RAW-C (Relatedness of Ambiguous Words, in Context). This resource contains both categorical annotations, derived from expert lexicographers (Merriam-Webster and the OED), as well as graded relatedness judgments from human participants. We found that contextualized representations from BERT and ELMo captured some variance ($R^2=.37$) in these relatedness judgments, but that the distinction between same-sense and different-sense uses, as well as between homonymy and polysemy, explains considerably more ($R^2 = .66$). Finally, we argued that this gap in performance represents an exciting opportunity for further development, and for cross-pollination between experimental psycholinguistics and NLP.

\section{Ethical Considerations}

All responses from human participants were anonymized before analyzing any data. Furthermore, the RAW-C dataset does not contain single-trial data---responses for a given sentence pair have been collapsed across all the human annotators who provided a rating for that pair. All participants provided informed consent, and were compensated in the form of SONA credits (to be applied to various Psychology, Cognitive Science, or Linguistics classes). The project was carried out with IRB approval.

\section*{Acknowledgments}

We are grateful to Susan Windisch Brown and Ekaterini Klepousniotou for making their experimental stimuli available. We also thank the anonymous reviewers for their helpful suggestions, and Nathan Schneider for early feedback on the idea to publish the dataset. Finally, we are grateful to other members of the Language and Cognition Lab (James Michaelov, Cameron Jones, and Tyler Chang) for valuable comments and discussion.


\bibliographystyle{acl_natbib}
\bibliography{acl2021}

\begin{thebibliography}{50}
\expandafter\ifx\csname natexlab\endcsname\relax\def\natexlab#1{#1}\fi

\bibitem[{Armendariz et~al.(2020)Armendariz, Purver, Pollak,
  Ljube{\v{s}}i{\'c}, Ul{\v{c}}ar, Vuli{\'c}, and
  Pilehvar}]{armendariz2020semeval}
Carlos~Santos Armendariz, Matthew Purver, Senja Pollak, Nikola
  Ljube{\v{s}}i{\'c}, Matej Ul{\v{c}}ar, Ivan Vuli{\'c}, and Mohammad~Taher
  Pilehvar. 2020.
\newblock \href {https://www.aclweb.org/anthology/2020.semeval-1.3/}
  {Sem{E}val-2020 {T}ask 3: Graded word similarity in context}.
\newblock In \emph{Proceedings of the Fourteenth Workshop on Semantic
  Evaluation}, pages 36--49.

\bibitem[{Bates et~al.(2015)Bates, M{\"a}chler, Bolker, and
  Walker}]{bates2007lme4}
Douglas Bates, Martin M{\"a}chler, Ben Bolker, and Steve Walker. 2015.
\newblock \href {https://doi.org/10.18637/jss.v067.i01} {Fitting linear
  mixed-effects models using {lme4}}.
\newblock \emph{Journal of Statistical Software}, 67(1):1--48.

\bibitem[{Bender and Koller(2020)}]{bender2020climbing}
Emily~M. Bender and Alexander Koller. 2020.
\newblock \href {https://www.aclweb.org/anthology/2020.acl-main.463/} {Climbing
  towards {N}{L}{U}: On meaning, form, and understanding in the age of data}.
\newblock In \emph{Proceedings of the 58th Annual Meeting of the Association
  for Computational Linguistics}, pages 5185--5198.

\bibitem[{Binder and Rayner(1998)}]{binder1998contextual}
Katherine~S Binder and Keith Rayner. 1998.
\newblock \href {https://link.springer.com/article/10.3758/BF03212950}
  {Contextual strength does not modulate the subordinate bias effect: Evidence
  from eye fixations and self-paced reading}.
\newblock \emph{Psychonomic Bulletin \& Review}, 5(2):271--276.

\bibitem[{Bisk et~al.(2020)Bisk, Holtzman, Thomason, Andreas, Bengio, Chai,
  Lapata, Lazaridou, May, Nisnevich et~al.}]{bisk2020experience}
Yonatan Bisk, Ari Holtzman, Jesse Thomason, Jacob Andreas, Yoshua Bengio, Joyce
  Chai, Mirella Lapata, Angeliki Lazaridou, Jonathan May, Aleksandr Nisnevich,
  et~al. 2020.
\newblock \href {https://arxiv.org/abs/2004.10151} {Experience grounds
  language}.
\newblock \emph{arXiv preprint arXiv:2004.10151}.

\bibitem[{Boleda et~al.(2019)Boleda, Gulordava, and Aina}]{boleda2019putting}
Gemma Boleda, Kristina Gulordava, and Laura Aina. 2019.
\newblock \href {https://arxiv.org/abs/1906.05149} {Putting words in context:
  {L}{S}{T}{M} language models and lexical ambiguity}.
\newblock In \emph{Proceedings of the 57th Annual Meeting of the Association
  for Computational Linguistics; 2019 Jul 28-Aug 2; Florence, Italy.
  Stroudsburg (PA): ACL; 2019. p. 3342--8.} ACL (Association for Computational
  Linguistics).

\bibitem[{Brown(2008{\natexlab{a}})}]{brown2008choosing}
Susan~Windisch Brown. 2008{\natexlab{a}}.
\newblock \href {https://www.aclweb.org/anthology/P08-2063.pdf} {Choosing sense
  distinctions for {W}{S}{D}: Psycholinguistic evidence}.
\newblock In \emph{Proceedings of ACL-08: HLT, Short Papers}, pages 249--252.

\bibitem[{Brown(2008{\natexlab{b}})}]{brown2008polysemy}
Susan~Windisch Brown. 2008{\natexlab{b}}.
\newblock \href {https://journals.colorado.edu/index.php/cril/article/view/289}
  {Polysemy in the mental lexicon}.
\newblock \emph{Colorado Research in Linguistics}, 21.

\bibitem[{Bruni et~al.(2014)Bruni, Tran, and Baroni}]{bruni2014multimodal}
Elia Bruni, Nam-Khanh Tran, and Marco Baroni. 2014.
\newblock \href {https://www.jair.org/index.php/jair/article/view/10857}
  {Multimodal distributional semantics}.
\newblock \emph{Journal of Artificial Intelligence Research}, 49:1--47.

\bibitem[{Dautriche(2015)}]{dautriche2015weaving}
Isabelle Dautriche. 2015.
\newblock \href {https://www.theses.fr/2015USPCB112} {\emph{Weaving an
  ambiguous lexicon}}.
\newblock Ph.D. thesis, Sorbonne Paris Cit{\'e}.

\bibitem[{Devlin et~al.(2018)Devlin, Chang, Lee, and
  Toutanova}]{devlin2018bert}
Jacob Devlin, Ming-Wei Chang, Kenton Lee, and Kristina Toutanova. 2018.
\newblock \href {https://arxiv.org/abs/1810.04805} {B{E}{R}{T}: Pre-training of
  deep bidirectional transformers for language understanding}.
\newblock \emph{arXiv preprint arXiv:1810.04805}.

\bibitem[{Elman(2009)}]{elman2009meaning}
Jeffrey~L Elman. 2009.
\newblock \href
  {https://onlinelibrary.wiley.com/doi/full/10.1111/j.1551-6709.2009.01023.x}
  {On the meaning of words and dinosaur bones: Lexical knowledge without a
  lexicon}.
\newblock \emph{Cognitive science}, 33(4):547--582.

\bibitem[{Faruqui and Dyer(2015)}]{faruqui2015non}
Manaal Faruqui and Chris Dyer. 2015.
\newblock \href {https://arxiv.org/abs/1506.05230} {Non-distributional word
  vector representations}.
\newblock \emph{arXiv preprint arXiv:1506.05230}.

\bibitem[{Fellbaum(1998)}]{fellbaum1998}
Christine Fellbaum, editor. 1998.
\newblock \emph{WordNet: An Electronic Lexical Database}.
\newblock Cambridge, MA: MIT Press.

\bibitem[{Finkelstein et~al.(2001)Finkelstein, Gabrilovich, Matias, Rivlin,
  Solan, Wolfman, and Ruppin}]{finkelstein2001placing}
Lev Finkelstein, Evgeniy Gabrilovich, Yossi Matias, Ehud Rivlin, Zach Solan,
  Gadi Wolfman, and Eytan Ruppin. 2001.
\newblock \href {https://dl.acm.org/doi/10.1145/503104.503110} {Placing search
  in context: The concept revisited}.
\newblock In \emph{Proceedings of the 10th international conference on World
  Wide Web}, pages 406--414.

\bibitem[{Gardner et~al.(2017)Gardner, Grus, Neumann, Tafjord, Dasigi, Liu,
  Peters, Schmitz, and Zettlemoyer}]{Gardner2017AllenNLP}
Matt Gardner, Joel Grus, Mark Neumann, Oyvind Tafjord, Pradeep Dasigi,
  Nelson~F. Liu, Matthew Peters, Michael Schmitz, and Luke~S. Zettlemoyer.
  2017.
\newblock \href {http://arxiv.org/abs/arXiv:1803.07640} {Allennlp: A deep
  semantic natural language processing platform}.

\bibitem[{Geeraerts(1993)}]{geeraerts1993vagueness}
Dirk Geeraerts. 1993.
\newblock \href
  {https://www.degruyter.com/document/doi/10.1515/cogl.1993.4.3.223/html}
  {Vagueness's puzzles, polysemy's vagaries}.
\newblock \emph{Cognitive Linguistics (includes Cognitive Linguistic
  Bibliography)}, 4(3):223--272.

\bibitem[{Gerz et~al.(2016)Gerz, Vuli{\'c}, Hill, Reichart, and
  Korhonen}]{gerz2016simverb}
Daniela Gerz, Ivan Vuli{\'c}, Felix Hill, Roi Reichart, and Anna Korhonen.
  2016.
\newblock \href {https://arxiv.org/abs/1608.00869} {Simverb-3500: A large-scale
  evaluation set of verb similarity}.
\newblock \emph{arXiv preprint arXiv:1608.00869}.

\bibitem[{Haber and Poesio(2020{\natexlab{a}})}]{haber2020assessing}
Janosch Haber and Massimo Poesio. 2020{\natexlab{a}}.
\newblock \href {https://www.aclweb.org/anthology/2020.starsem-1.12/}
  {Assessing polyseme sense similarity through co-predication acceptability and
  contextualised embedding distance}.
\newblock In \emph{Proceedings of the Ninth Joint Conference on Lexical and
  Computational Semantics}, pages 114--124.

\bibitem[{Haber and Poesio(2020{\natexlab{b}})}]{haber2020word}
Janosch Haber and Massimo Poesio. 2020{\natexlab{b}}.
\newblock \href {https://www.aclweb.org/anthology/2020.pam-1.17/} {Word sense
  distance in human similarity judgements and contextualised word embeddings}.
\newblock In \emph{Proceedings of the Probability and Meaning Conference (PaM
  2020)}, pages 128--145.

\bibitem[{Halawi et~al.(2012)Halawi, Dror, Gabrilovich, and
  Koren}]{halawi2012large}
Guy Halawi, Gideon Dror, Evgeniy Gabrilovich, and Yehuda Koren. 2012.
\newblock \href
  {https://dl.acm.org/doi/abs/10.1145/2339530.2339751?casa_token=7SXjDWIA05MAAAAA:1k62Hf1Ovl-ySGDKQsJ0l1BSn79AyHRhXyjzEDSPbTnse1JNCaOsuukav3VDNm8BqCfZ_JLdQso}
  {Large-scale learning of word relatedness with constraints}.
\newblock In \emph{Proceedings of the 18th ACM SIGKDD international conference
  on Knowledge discovery and data mining}, pages 1406--1414.

\bibitem[{Hanks(2000)}]{hanks2000word}
Patrick Hanks. 2000.
\newblock \href {https://www.jstor.org/stable/30204810} {Do word meanings
  exist?}
\newblock \emph{Computers and the Humanities}, 34(1/2):205--215.

\bibitem[{Hill et~al.(2015)Hill, Reichart, and Korhonen}]{hill2015simlex}
Felix Hill, Roi Reichart, and Anna Korhonen. 2015.
\newblock \href {https://www.mitpressjournals.org/doi/abs/10.1162/COLI_a_00237}
  {Simlex-999: Evaluating semantic models with (genuine) similarity
  estimation}.
\newblock \emph{Computational Linguistics}, 41(4):665--695.

\bibitem[{Huang et~al.(2012)Huang, Socher, Manning, and
  Ng}]{huang2012improving}
Eric~H Huang, Richard Socher, Christopher~D Manning, and Andrew~Y Ng. 2012.
\newblock \href {https://www.aclweb.org/anthology/P12-1092.pdf} {Improving word
  representations via global context and multiple word prototypes}.
\newblock In \emph{Proceedings of the 50th Annual Meeting of the Association
  for Computational Linguistics (Volume 1: Long Papers)}, pages 873--882.

\bibitem[{Kilgarriff(2007)}]{kilgarriff2007word}
Adam Kilgarriff. 2007.
\newblock \href {https://link.springer.com/chapter/10.1007/978-1-4020-4809-8_2}
  {Word senses}.
\newblock In \emph{Word Sense Disambiguation}, pages 29--46. Springer.

\bibitem[{Klepousniotou(2002)}]{klepousniotou2002processing}
Ekaterini Klepousniotou. 2002.
\newblock \href
  {https://www.sciencedirect.com/science/article/abs/pii/S0093934X01925180}
  {The processing of lexical ambiguity: Homonymy and polysemy in the mental
  lexicon}.
\newblock \emph{Brain and language}, 81(1-3):205--223.

\bibitem[{Klepousniotou and Baum(2007)}]{klepousniotou2007disambiguating}
Ekaterini Klepousniotou and Shari~R Baum. 2007.
\newblock \href
  {https://www.sciencedirect.com/science/article/abs/pii/S0911604406000145}
  {Disambiguating the ambiguity advantage effect in word recognition: An
  advantage for polysemous but not homonymous words}.
\newblock \emph{Journal of Neurolinguistics}, 20(1):1--24.

\bibitem[{Klepousniotou et~al.(2008)Klepousniotou, Titone, and
  Romero}]{klepousniotou2008making}
Ekaterini Klepousniotou, Debra Titone, and Carolina Romero. 2008.
\newblock \href {https://pubmed.ncbi.nlm.nih.gov/18980412/} {Making sense of
  word senses: The comprehension of polysemy depends on sense overlap.}
\newblock \emph{Journal of Experimental Psychology: Learning, Memory, and
  Cognition}, 34(6):1534.

\bibitem[{Lacerra et~al.(2020)Lacerra, Bevilacqua, Pasini, and
  Navigli}]{lacerra2020csi}
Caterina Lacerra, Michele Bevilacqua, Tommaso Pasini, and Roberto Navigli.
  2020.
\newblock \href {https://ojs.aaai.org/index.php/AAAI/article/view/6324}
  {{C}{S}{I}: A coarse sense inventory for 85\% word sense disambiguation.}
\newblock In \emph{AAAI}, pages 8123--8130.

\bibitem[{Leinenger and Rayner(2013)}]{leinenger2013eye}
Mallorie Leinenger and Keith Rayner. 2013.
\newblock \href
  {https://www.tandfonline.com/doi/full/10.1080/20445911.2013.806513} {Eye
  movements while reading biased homographs: Effects of prior encounter and
  biasing context on reducing the subordinate bias effect}.
\newblock \emph{Journal of Cognitive Psychology}, 25(6):665--681.

\bibitem[{Levine et~al.(2020)Levine, Lenz, Dagan, Ram, Padnos, Sharir,
  Shalev-Shwartz, Shashua, and Shoham}]{levine-etal-2020-sensebert}
Yoav Levine, Barak Lenz, Or~Dagan, Ori Ram, Dan Padnos, Or~Sharir, Shai
  Shalev-Shwartz, Amnon Shashua, and Yoav Shoham. 2020.
\newblock \href {https://www.aclweb.org/anthology/2020.acl-main.423}
  {{S}ense{BERT}: Driving some sense into {BERT}}.
\newblock In \emph{Proceedings of the 58th Annual Meeting of the Association
  for Computational Linguistics}, pages 4656--4667, Online. Association for
  Computational Linguistics.

\bibitem[{Li and Joanisse(2021)}]{li2021word}
Jiangtian Li and Marc~F Joanisse. 2021.
\newblock \href {https://onlinelibrary.wiley.com/doi/10.1111/cogs.12955?af=R}
  {Word senses as clusters of meaning modulations: A computational model of
  polysemy}.
\newblock \emph{Cognitive Science}, 45(4):e12955.

\bibitem[{Lopukhina et~al.(2018)Lopukhina, Laurinavichyute, Lopukhin, and
  Dragoy}]{lopukhina2018mental}
Anastasiya Lopukhina, Anna Laurinavichyute, Konstantin Lopukhin, and Olga
  Dragoy. 2018.
\newblock \href
  {https://www.frontiersin.org/articles/10.3389/fpsyg.2018.00192/full} {The
  mental representation of polysemy across word classes}.
\newblock \emph{Frontiers in psychology}, 9:192.

\bibitem[{Loureiro et~al.(2020)Loureiro, Rezaee, Pilehvar, and
  Camacho-Collados}]{loureiro2020language}
Daniel Loureiro, Kiamehr Rezaee, Mohammad~Taher Pilehvar, and Jose
  Camacho-Collados. 2020.
\newblock \href {https://arxiv.org/abs/2008.11608} {Language models and word
  sense disambiguation: An overview and analysis}.
\newblock \emph{arXiv preprint arXiv:2008.11608}.

\bibitem[{Miller et~al.(1993)Miller, Leacock, Tengi, and
  Bunker}]{miller1993semantic}
George~A Miller, Claudia Leacock, Randee Tengi, and Ross~T Bunker. 1993.
\newblock A semantic concordance.
\newblock In \emph{Human Language Technology: Proceedings of a Workshop Held at
  Plainsboro, New Jersey, March 21-24, 1993}.

\bibitem[{Nair et~al.(2020)Nair, Srinivasan, and
  Meylan}]{nair2020contextualized}
Sathvik Nair, Mahesh Srinivasan, and Stephan Meylan. 2020.
\newblock \href {https://arxiv.org/abs/2010.13057} {Contextualized word
  embeddings encode aspects of human-like word sense knowledge}.
\newblock \emph{arXiv preprint arXiv:2010.13057}.

\bibitem[{Pennington et~al.(2014)Pennington, Socher, and
  Manning}]{pennington2014glove}
Jeffrey Pennington, Richard Socher, and Christopher~D Manning. 2014.
\newblock \href {https://www.aclweb.org/anthology/D14-1162.pdf} {Glo{V}e:
  Global vectors for word representation}.
\newblock In \emph{Proceedings of the 2014 conference on empirical methods in
  natural language processing (EMNLP)}, pages 1532--1543.

\bibitem[{Peters et~al.(2018)Peters, Neumann, Iyyer, Gardner, Clark, Lee, and
  Zettlemoyer}]{peters2018deep}
Matthew~E Peters, Mark Neumann, Mohit Iyyer, Matt Gardner, Christopher Clark,
  Kenton Lee, and Luke Zettlemoyer. 2018.
\newblock \href {https://arxiv.org/abs/1802.05365} {Deep contextualized word
  representations}.
\newblock \emph{arXiv preprint arXiv:1802.05365}.

\bibitem[{Pilehvar and Camacho-Collados(2018)}]{pilehvar2018wic}
Mohammad~Taher Pilehvar and Jose Camacho-Collados. 2018.
\newblock \href {https://arxiv.org/abs/1808.09121} {Wi{C}: the word-in-context
  dataset for evaluating context-sensitive meaning representations}.
\newblock \emph{arXiv preprint arXiv:1808.09121}.

\bibitem[{Rayner et~al.(1994)Rayner, Pacht, and Duffy}]{rayner1994effects}
Keith Rayner, Jeremy~M Pacht, and Susan~A Duffy. 1994.
\newblock \href
  {https://www.sciencedirect.com/science/article/abs/pii/S0749596X84710254}
  {Effects of prior encounter and global discourse bias on the processing of
  lexically ambiguous words: Evidence from eye fixations}.
\newblock \emph{Journal of memory and language}, 33(4):527--544.

\bibitem[{Rodd et~al.(2004)Rodd, Gaskell, and
  Marslen-Wilson}]{rodd2004modelling}
Jennifer~M Rodd, M~Gareth Gaskell, and William~D Marslen-Wilson. 2004.
\newblock \href
  {https://onlinelibrary.wiley.com/doi/abs/10.1207/s15516709cog2801_4}
  {Modelling the effects of semantic ambiguity in word recognition}.
\newblock \emph{Cognitive science}, 28(1):89--104.

\bibitem[{Srinivasan and Rabagliati(2015)}]{srinivasan2015concepts}
Mahesh Srinivasan and Hugh Rabagliati. 2015.
\newblock \href
  {https://www.sciencedirect.com/science/article/pii/S0024384114002885} {How
  concepts and conventions structure the lexicon: Cross-linguistic evidence
  from polysemy}.
\newblock \emph{Lingua}, 157:124--152.

\bibitem[{Taieb et~al.(2020)Taieb, Zesch, and Aouicha}]{taieb2020survey}
Mohamed Ali~Hadj Taieb, Torsten Zesch, and Mohamed~Ben Aouicha. 2020.
\newblock \href {https://link.springer.com/article/10.1007/s10462-019-09796-3}
  {A survey of semantic relatedness evaluation datasets and procedures}.
\newblock \emph{Artificial Intelligence Review}, 53(6):4407--4448.

\bibitem[{Trott et~al.(2020)Trott, Torrent, Chang, and Schneider}]{trott2020re}
Sean Trott, Tiago~Timponi Torrent, Nancy Chang, and Nathan Schneider. 2020.
\newblock \href {https://arxiv.org/pdf/2005.09099.pdf} {{(Re)} construing
  {M}eaning in {N}{L}{P}}.
\newblock In \emph{Proceedings of the 58th Annual Meeting of the Association
  for Computational Linguistics (ACL 2020)}.

\bibitem[{Tuggy(1993)}]{tuggy1993ambiguity}
David Tuggy. 1993.
\newblock \href
  {https://www.degruyter.com/document/doi/10.1515/cogl.1993.4.3.273/html}
  {Ambiguity, polysemy, and vagueness}.
\newblock \emph{Cognitive linguistics}, 4(3):273--290.

\bibitem[{Valera(2020)}]{valera2020polysemy}
Salvador Valera. 2020.
\newblock Polysemy versus homonymy.
\newblock In \emph{Oxford Research Encyclopedia of Linguistics}.

\bibitem[{Wiedemann et~al.(2019)Wiedemann, Remus, Chawla, and
  Biemann}]{wiedemann2019does}
Gregor Wiedemann, Steffen Remus, Avi Chawla, and Chris Biemann. 2019.
\newblock \href {https://arxiv.org/abs/1909.10430} {Does {B}{E}{R}{T} make any
  sense? {I}nterpretable {W}ord {S}ense {D}isambiguation with contextualized
  embeddings}.
\newblock \emph{arXiv preprint arXiv:1909.10430}.

\bibitem[{Yang et~al.(2019)Yang, Dai, Yang, Carbonell, Salakhutdinov, and
  Le}]{yang2019xlnet}
Zhilin Yang, Zihang Dai, Yiming Yang, Jaime Carbonell, Russ~R Salakhutdinov,
  and Quoc~V Le. 2019.
\newblock \href {https://arxiv.org/abs/1906.08237} {Xlnet: Generalized
  autoregressive pretraining for language understanding}.
\newblock In \emph{Advances in neural information processing systems}, pages
  5753--5763.

\bibitem[{Yee and Thompson-Schill(2016)}]{yee2016putting}
Eiling Yee and Sharon~L Thompson-Schill. 2016.
\newblock \href {https://link.springer.com/article/10.3758/s13423-015-0948-7}
  {Putting concepts into context}.
\newblock \emph{Psychonomic bulletin \& review}, 23(4):1015--1027.

\bibitem[{Yurchenko et~al.(2020)Yurchenko, Lopukhina, and
  Dragoy}]{yurchenko2020metaphor}
Anna Yurchenko, Anastasiya Lopukhina, and Olga Dragoy. 2020.
\newblock \href
  {https://www.frontiersin.org/articles/10.3389/fpsyg.2020.02113/full?report=reader}
  {Metaphor is between metonymy and homonymy: Evidence from event-related
  potentials}.
\newblock \emph{Frontiers in Psychology}, 11:2113.

\end{thebibliography}


\end{document}